# Vezetői modellek tipizálása klaszterezési módszerek segítségével

# Typification of Driver Models Using Clustering Methods

**Ignéczi Gergő Ferenc[a], Dobay Tamás[b]**

[a]Széchenyi István Egyetem, Járműipari Kutatóközpont
gergo.igneczi@ga.sze.hu

[b]Széchenyi István Egyetem, Járműipari Kutatóközpont
dobayt@gmail.com

**Absztrakt:** Az automatizált vezetési rendszerek rohamos fejlődése az elmúlt években a közlekedésbiztonság és az utazási kényelem javulásához vezetett. Az egyik jellemző funkciója ezeknek a rendszereknek a sávtartás, ami általános esetben nem veszi figyelembe az emberi vezetésre jellemző preferenciákat. Korábbi munkánk során bebizonyítottuk, hogy egy lineáris vezetői modellen alapuló útvonal tervező algoritmussal lehetséges az emberi preferenciáknak megfelelő sávtartó funkciót megvalósítani. Jelenlegi munkánk során azt vizsgáljuk, miként lehet külön választani az egyes sofőrökre jellemző vezetési stílusokat az ívválasztási preferenciájuk alapján. Ennek bizonyítására klaszterezési módszereket alkalmaztunk korábban rögzített méréseken. Az osztályokhoz tartozó átlagos viselkedést leíró paraméterekkel szimulációkat végezve (klaszterezett típusokkal újra szimulálva) kiderült, hogy az így kapott útvonalak alapján a sofőrök osztályozása sikeres volt, a 3 osztály viselkedésében elkülönül és a modellünk segítségével ezen viselkedések reprodukálhatók.

**Kulcsszavak**: vezetői modell, klaszterezés, validáció

**Abstract:** The rapid development of automated driving systems in recent years has led to improvements in road safety and travel comfort. One typical function of these systems is Lane Keep Assist, which generally does not take human driving preferences into account. In our previous work, we have demonstrated that it is possible to implement a Lane Keep Assist function that is appropriate to human preferences using a trajectory planning algorithm based on a linear driving model. In our current work, we investigated how to separate the driving styles of individual drivers. We assumed that there are three driving styles: sporty, neutral and defensive. To prove these relations, clustering methods were applied to previously recorded measurements . Simulations with parameters describing the average behaviour of the classes (re-simulated with clustered types) showed that the resulting paths successfully classified drivers, that the 3 classes are distinct in their behaviour and that our model reproduces these behaviours.

## 1. Bevezetés

A 2020-as évek autóipari trendjei két fő irány köré rendeződnek: energiamenedzsment és automatizálás. A járművezetés automatizálása évtizedes ötlet, már az 1980-as és 90-es években jelentek meg olyan asszitens rendszerek, melyek segítették a vezető feladatát ezzel növelve a vezetés biztonságát és komfortélményét. Az első ilyen rendszerek voltak a blokkolásgátló rendszerek (Anti-Blocking Systems, ABS), amelyet a Chrysler és a Mercedes mutatott be az 1970-es években. Ennek folyománya a kipörgésgátló (Traction Control, TC), illetve az Elektronikus Stabilitás Kontrol (Electronic Stability Control, ESC), amelyet több japán, német, illetve amerikai gyártó is korán kifejlesztett, már az 1990-es években. Ezek elsősorban a vezetés biztonságát növelték, így mára már az összes újonnan üzembe helyezett járműben kötelező a jelenlétük. Ezzel párhuzamosan megjelentek az ún. kényelmi funkciók, amelyek nem kizárólag a biztonságot növelik, hanem vezetési feladatokat vesznek át a sofőrtől, ezzel könnyítve a járművezetést. A jelenleg elterjedt rendszerek hosszirányú, illetve keresztirányú járműirányítási feladatokra bomlanak. A legelterjedtebb hosszirányú szabályzást megvalósító funkció az adaptív sebességtartó automatika (Adaptive Cruise Control, ACC) [1]. Az ACC képes a jármű sebességének szabályzására a teljes működési tartományon, akár állóhelyzetből a maximális sebességig úgy, hogy figyelembe veszi más járművek sebességét, illetve a sebességhatárokat, vagy a kanyarívből adódó sebességcsökkentést. A legelterjedtebb keresztirányú funkció, mely kényelmi célokat szolgál az ún. Sávtartó Automatika (Lane Keep System, LKS) [2]. Ennek célja, hogy a járművet az úton, illetve a sávon belül tartsa úgy, hogy biztonságos távolságot hagyjon más objektumoktól. Más elnevezése ennek a rendszernek az ún. Lane Centering System, vagyis „sávközépre szabályzó rendszer". Ez az elnevezés már a technológiai megvalósításra is utal, hiszen a legtöbbször a járművet a sávtartó automatika a sáv közepén vezeti, ezzel biztosítva a fentebb említett biztonsági követelményeket. Annak ellenére, hogy ez a megközelítés technológiailag egy jó megoldás, erősen gépies, hiszen az emberek legtöbbször nem a sáv közepén vezetnek.

A fentebb említett asszisztens rendszerek fejlődésével megjelent az igény a jármű minél nagyobb önállóságára, arra, hogy a jármű minél nagyobb tartományban tudjon önállóan vezetni, emberi kooperációval és/vagy felügyelettel. Ezen automatizáltsági szintek osztályozására a Society of Automotive Engineers (SAE) 2014-ben egységes rendszert vezetett be [3]. Összesen 5 szintet különböztetünk meg. Az 1. automatizáltsági szint jelenti a legkisebb automatizáltságot, ahol az emberi vezető teljes mértékben jelen van és a felelősség a vezetésért is őt terheli. Azonban bizonyos feladatokban a jármű átveszi az irányítást, ilyen pl. a sebességtartó automatika (tempomat). A 2. és 3. szinten egyre inkább nagyobb hangsúly helyeződik a járműre, a jármű által ellátott feladatok mennyisége és azok széles működési tartományára, ugyanakkor a sofőrnek jelen kell lennie, komplikációk esetén az ő feladata az irányítás átvétele. A 4. és 5. szinten a járművezető szerepe szinte nullára csökken. Az 1. szintet szokás asszisztenciának, a 2. és 3. szint automatikának és a 4. illetve 5. szintet autonómiának, azaz önvezetésnek hívni. Megjegyezzük, hogy a határok nem élesek, így a szintek között van átfedés, ennek megfelelően az elnevezések is változhatnak (pl. a fejlesztés egységesítéséért német autógyártók bevezettek egy ún. 2+ szintet, ezzel némileg teret adva a fejlesztéseknek, amíg a 3. szinthez szükséges jogi háttér nem születik meg). Ahogy nő az automatizáltsági szint, úgy fogalmazódnak meg további felhasználói igények a szóban forgó rendszerek viselkedésével kapcsolatban. A vezetési feladat átadása nagyfokú bizalmat igényel. A felhasználók bizalma az automatizált vezetési rendszerekben kulcskérdés ahhoz, hogy ezek a rendszerek elterjednek. Minél inkább kielégítjük a felhasználó preferenciáit a működés jellegével, annál inkább válik elfogadottá egy automatizált vezetési rendszer. A 2020-as évek elején ez az igény már az autógyártók részéről is felmerült, így kulcskérdéssé vált: hogyan tudjuk az automatizált vezetési rendszereket személyre szabhatóvá, emberszerűvé tenni [4] [5] [6] [7]? Jelen cikkünk célja, hogy a 2. és 3. szintű sávtartó automatikák viselkedése emberszerűbbé váljon, illetve megoldásokat keresünk arra, hogyan lehet manuális vezetés alatt megismerni a vezetői egyéni preferenciáit. Ezért felállítottunk egy vezetői modellt, mely képes reprodukálni emberszerű útvonalakat a sávban történő haladáshoz. Ebben a cikkben bemutatjuk, hogyan hasznosítható ez a modell a fenti alkalmazásban, illetve beazonosítunk több vezetői típust a modell segítségével.

A cikk felépítése a következő: először bemutatjuk a vezetői modellezés irodalmát, illetve egy korábbi kutatásunk során azonosított vezetői modellt, amelyet felhasználunk az ívválasztási preferenciák azonosítására. Ezek után bemutatjuk a modell alkalmazását, a modell paraméterek alapján a vezetési preferenciák klaszterezésének lehetőségeit. Végezetül egy átfogó validációt ismertetünk, amely során bebizonyítjuk, hogy a feltárt vezetői klaszterek milyen vezetői típusokat jelentenek, illetve ezen típusoknak milyen jellemzői vannak.

## 2. Irodalmi áttekintés

### a. A vezetői modellek csoportosítása

A vezetési feladatok, és ezzel együtt a vezető, mint ember modellezése több évtizedre tekint vissza. Már az 1950-es években vizsgálták autóvezetők követési távolságát, és az ehhez kapcsolódó viselkedési normákat [4]. Az ezt követő évtizedekben a vezetési modellek kiegészültek, és a forgalom egészét kezdték vizsgálni [5] [6]. Ennek elsődleges célja az volt, hogy megértsék az emberi sofőrök viselkedését és ezáltal fejlesszék az útbiztonságot. A modellek ebben az esetben kifejezetten leíró szerepet töltöttek be, így jóslatokat tettek egy-egy útszakaszon várható balesetek számára, vagy épp mikroszkopikus forgalmi helyzetek alakulására. Ezen kívül a modellek segítettek megérteni, hogyan fejlődik egy járművezető, ezáltal jobb vezetői képzést tudtak nekik nyújtani.

Az első jelentős összefoglalását a vezetői modellezési feladatoknak 1984-ben Evans és Schwing tette meg, melyet később John A. Michon holland mérnök dolgozott át [7]. Ebben a sokat hivatkozott tanulmányban szerepel minden jelentős 1950 és 1985 közötti vezetői modell. A teljes közlekedési rendszert Evans és Schwing a 1. táblázat szerinti szempontok mentén foglalta össze. Eszerint a közlekedési rendszerben való részvétel felosztható az egyén két fő stratégiájára:
- osztályozási stratégia, azaz a döntéshozáshoz szükséges egyszerűsítés, információ észlelés és feldolgozás, majd döntéshozás, mindezt egyéni preferenciák alapján, illetve
- funkcionális stratégia, azaz a döntés alapján megvalósítás, cselekvés, végrehajtás.

A másik dimenziót képezik a modellek, amelyek megtestesítik ezeket a stratégiákat feloszthatók az alábbi két csoportra:
- ki-, és bemeneti modellek, azaz olyan viselkedési modellek, amelyek valamilyen bemeneti adat (érzékelés) alapján produkálnak kimenetet (pl. cselekvést, származtatott információt vagy döntést), illetve
- belső állapotok, amelyek pszichológiai modellek és az egyén személyes preferenciái alapján megadják a jellemző viselkedést rövid- és hosszútávon.

|  | Osztályozási stratégia | Funkcionális stratégia |
|---|---|---|
| Ki – Bemeneti modellek (viselkedési modellek) | Feladatanalízis | Mechanikai modellek<br>Adaptív Szabályzó Modellek:<br>- szervó-szabályzás<br>- információ áramlás szabályzása |
| Belső állapotok (pszichológiai modellek) | Tulajdonság modellek | Motivációs modellek<br>Kognitív (leíró) modellek |

**1. táblázat.** Evans és Schwing közlekedési rendszer modellje, az eredeti táblázat forrása: **[7]**

A fenti két dimenzió alapján összesen négy nagyobb feladatcsoportot különítünk el: feladatanalízis, tulajdonság modellek, funkcionális ki- és bementi modellek és motivációs vagy leíró modellek.

A feladatanalízist leíró modellek célja, hogy elkülönítse a vezetés során felmerülő feladatokat, illetve megadja azon összefüggéseket, amelyek segítségével mérhető mennyiségek alapján a feladatok elkülöníthetők, osztályozhatók. Egy ilyen rendszert dolgozott ki McKnight és Adams [6]. Összesen 45 főbb feladatot különítettek el. Ezek reprezentálják a vezetés során felmerülő feladatokat, viselkedési célokat, illetve az egyéni és a körülményekből adódó képességeket. Ide sorolhatók azok a leírások is, amelyek kifejezetten a vezetés során felmerülő hibalehetőségeket vizsgálják.

A tulajdonság modellek tartalmazzák az egyéni tulajdonságokat, melyek szándékolt jellemzők és befolyásolják a vezetési feladathoz való hozzáállást (pl. személyiségtípus, hangulat). Ezen kívül ide sorolhatjuk a vezetési képesség hiányából adódó nem megfelelő normákat. Ezen a területen érdemes megemlíteni pl. a kifejezetten motoros tevékenységek fejlődésére irányuló kutatásokat, melyek hozzájárulnak a vezetői oktatás javításához, a XXI. században pedig a vezetéstámogató rendszerek fejlesztéséhez [5].

A funkcionális ki- és bementi modellek a legelterjedtebbek az ADAS (Automated Driving and Assisted System) területén. Ezen kívül a forgalom modellezésére is előszeretettel használják. Ide sorolhatók az 1980-as évektől fejlesztett vezetési modellek mind motoros szinten [8], [9] mind jármű szinten [10], [11], [12], [13], [14], [15].

A funkcionális, ám elsősorban belső állapotokra visszavezethető modellek jellemzően a tervezés szintjén jelennek meg. Ez azt jelenti, hogy a vezetői valamilyen szempontrendszer szerint kiválasztja a végrehajtandó mozgásbéli feladatot, majd a motoros tevékenységein keresztül ezt meg is valósítja. Az 1985-ös Michon cikkig ezen a szinten elsősorban olyan munkák jelentek meg, melyek az ember toleranciáját, azaz a jármű mozgásában bekövetkező hibákat, illetve a vezető ezen hibákhoz való hozzáállását vizsgálja. Ide kapcsolódnak a kompenzációs modellek, a kockázat adaptációs elmélet („*risk homeostatis theory*") [16], a kockázat határérték („*risk threshold theory*") elmélete [17], illetve a fenyegetés elkerülő modellek („*threat avoidance models*") [18]. Ezek a modellek kifejezetten hasznosak akkor, ha a modellezés célja a leírás, azaz egy megfigyelt jelenség (mért adat) magyarázása. ADAS rendszerekkel kapcsolatos alkalmazásokban ezen modellek által leírt viselkedéseket általában nem reprodukáljuk, hiszen az ember tűréshatárán belüli hibáit nem szeretnénk viszont látni egy gépi irányítástól. Ugyanakkor jó alapot adhat a követelményeink megalkotására, pl. milyen hibahatárt engedünk meg egy járműszabályzásnak. Emellett ezen modell típusokba sorolhatók az ADAS felhasználásban is közkedvelt vezetői modellek, melyek a befutni kívánt útvonalat adják meg. Ezek a modellek szolgálnak alapul a jármű kívánt mozgásához, amely esetén szeretnénk emberszerűségre törekedni [19], [20], [21], [22], [23], [24], [25], [26], [27], [28], [29], [30], [31].

### b. Vezetői modell rendszerek

Evans és Schwing modell osztályozása egy jó összegzése a modellezési feladatoknak, de nem ad konkrét megoldást a modellezési problémára. Ugyanakkor ők is hivatkoznak olyan megközelítésekre, amelyek alkalmazhatók a gyakorlatban. A vezető feladatainak Jannsen-féle hierarchikus modelljét mutatja az 1. ábra. Ugyanezen modellen alapszanak későbbi munkák, melyek ADAS architektúrákat ajánlanak [32], [33]. Ez a modell összesen 3 hierarchikus szintet tartalmaz:

- stratégiai szint, amely kifejezetten az általános tervezést, a leghosszabb időhorizonton való döntéshozást tartalmazza (pl. két város közötti útvonal kiválasztása),
- manőver szint, amelyet sokszor taktikai szintnek is hívunk, amely kifejezetten az aktuális információk alapján hozott döntés (tervezés), figyelembe véve a személyes preferenciákat (pl. ívválasztás, sebességválasztás, gyorsítás mértéke, sávváltás melletti döntés…stb.), illetve
- szabályzó szint, amelyet operatív szintnek is hívunk, és az eltervezett viselkedés megvalósításáért felel, a legtöbbször tudatalatti, motoros tevékenységek egymásutáni végrehajtása segítségével (pl. a gázpedál vagy kormánykerék kezelése).

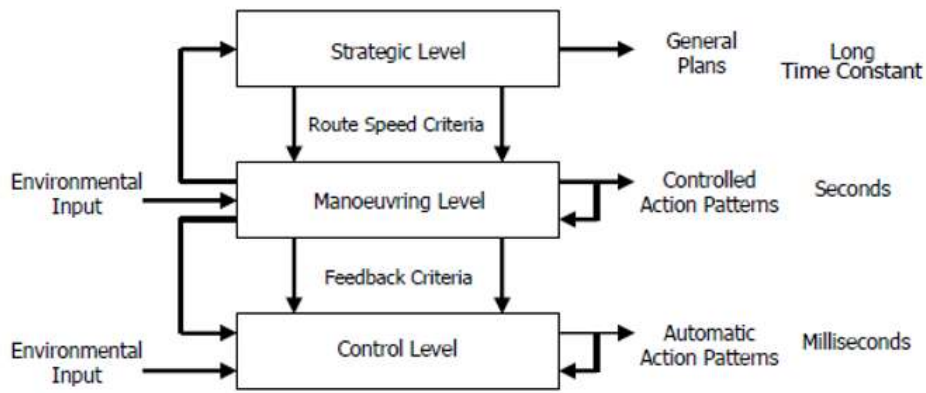

**1. ábra. A vezető feladatainak hierarchikus modellje, forrás: [7]**

Ez a 3 szint a feladatok egymáshoz való kapcsolódását modellezi. Egy másik dimenzió szerint rendezhetjük a vezetési feladatszinteket az egyes szintek célja szerint. Erre ad példát D.C. Wickens az információ áramlási modelljével, amely leginkább az 1. táblázat szerinti funkcionális ki- és bemeneti modelljei közé sorolható [34]. A modellt a 2. ábra mutatja. Ez alapján három fő információs szint különíthető el:
- észlelés és érzékelés („*perception and detection*"),
- döntéshozás („*decision making*"),
- végrehajtás („*execution*").

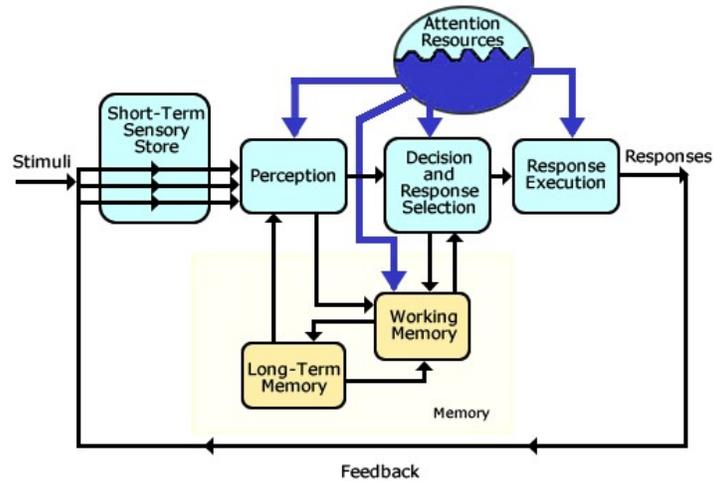

**2. ábra. D. C. Wickens információ áramlási modellje, forrás: [34]**

A három információs szint mindegyike kapcsolódik a rövid és hosszútávú memóriákon keresztül, és ezzel születik meg a végső beavatkozás.

Látható, hogy a vezetői modellezés több dimenzióban vizsgálandó. Elkülönül az információ időbeli feldolgozása, illetve a feladatok szintje. Ezt a többdimenziós megközelítést használva született meg a vezetői modellek 3-dimenziós struktúrája, melyet a 3. ábra mutat. J. Theeuwes modelljében megjelenik mind az információs, mind a feladathierarchikus modell, kiegészülve az ún. feladat teljesítés („*task performance*") modelljével. Ez utóbbi elkülönít 3 különböző módot:
- készségalapú,
- szabályalapú,
- tudásalapú feladatvégrehajtás.

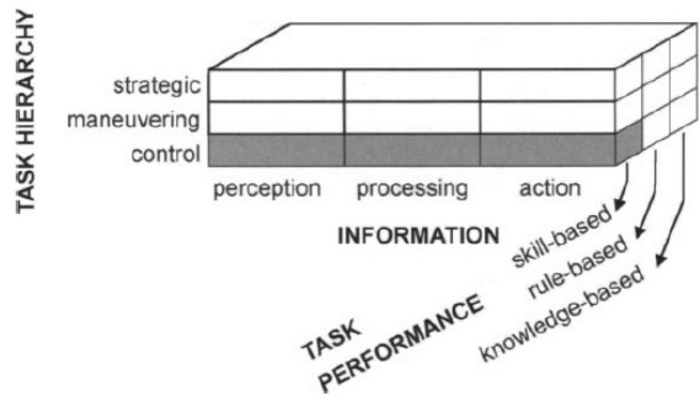

**3. ábra. A 3 dimenziós vezetői modellrendszer szemléltetése, forrás: [35]**

Theeuwes modelljéhez hasonló motiváció búik meg a GADGET projekt által ajánlott vezetői modellben [36]. Ez a modell Jannsen hiearchikus modelljén alapszik, kiegészítve azt egy 4., ún. viselkedési szinttel. A kiegészített modellt a 4. ábra mutatja. A viselkedési szint tartalmazza mindazon egyéni tulajdonságokat, amelyek befolyásolják az alatta lévő szintek feladatait. Theeuwes háromdimenziós modelljével ez a megközelítés hasonlóságot mutat, hiszen a feladat teljesítés módja függ az egyéni tulajdonságoktól (pl. egy kezdő sofőr a jármű irányítási szinten inkább tudásalapon működik, egy tapasztalt sofőr inkább készség alapon).

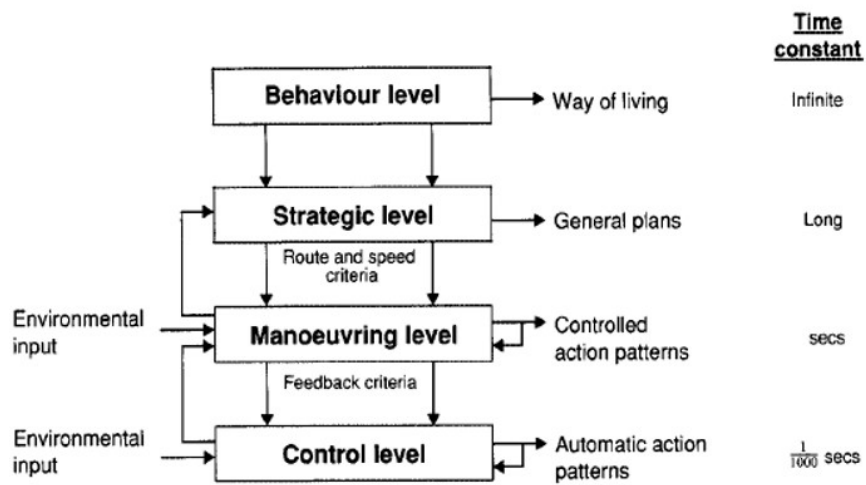

4. ábra. A GADGET projekt által ajánlott kiegészített hierarchikus modell, forrás: [36]

### c. Vezetői modellek irodalmának összegzése

Az eddigi fejezetekben áttekintést adtunk a vezetői modellek XX. századi fejlődéséről, irodalmáról. Habár a vezetés modellezése komplex feladat, mely a pszichológia és a mérnöki tudományok határán helyezkedik el, megkíséreljük az alábbiak szerint egyszerűen összefoglalni a következtetéseket:
- a vezetői modellek csoportosíthatók a modellek céljai és megoldásai alapján, lásd 1. táblázat,
- a vezetői modellrendszerek több osztályba is sorolhatók (pl. Jannsen hiearchikus modellje, Theeuwes háromdimenziós modellje, Wickens információ áramlási modellje): ezekben megjelenik mind az osztályozási feladat (pl. a döntéshozási szinten, stratégiai szinten…stb.), mind a funkcionális feladat (végrehajtás, irányítás és manőverezés),
- a vezetői modellek minimum 4 dimenzióra bonthatók: feladathierarchia, információ áramlás szintjei, feladat teljesítés módja és viselkedési modell.

Az érthetőség kedvéért a következő fogalmakat vezetjük be:

- modell-rendszer: a vezetői modellek legnagyobb, összefoglaló szintje, amely több dimenzióban reprezentálja a vezetői feladatokat,
- modell-alrendszer: a modell rendszer dimenziója, több modell összessége,
- modell: a vezetői modell egysége, pl. a feladathierarchikus modellben a szabályzó szint modellje; ezen szint a modell alrendszer több eleméhez is kapcsolódhat (pl. a 3-dimenziós modell-rendszer egy építőköve),
- almodellek: a modelleket felépítő matematikai modellek.

### 3. A modellezési probléma

#### a. A probléma megfogalmazása

Az eddigiekben a vezetői modellek osztályozásáról szóltunk. Ahhoz, hogy a jelen cikkben tárgyalt modellezési problémát megfogalmazzuk, röviden kitérünk a modell céljára, illetve annak felhasználási területére. A vezetési rendszer modellezése több céllal is történhet. [37] szerint egy vezető modellezése történhet abból a célból, hogy a megfigyelt vezetési mintákat meg tudjuk magyarázni, vagy hogy elkülönítsük a biztonságos és nem biztonságos vezetési attitűdöket. Ezen kívül a 2000-es évek elejétől nagy hangsúlyt kap a vezetői modellek asszisztens és automatizált rendszerekben történő felhasználása. Ez utóbbi jól megfigyelhető az irodalomban: 2000 után a vezetői modellek elsőszámú alkalmazása az ún. ADAS területen történik [38], [39], [40], [41], [8], [9], [42]. Ahhoz azonban, hogy ezek az alkalmazások a megfelelő szintű vezetői modellt használják, tisztában kell lenni a teljes közlekedési rendszer modelljével, aminek egy része a vezető maga. Ugyanerre hívja fel a figyelmet Cody és Gordon: *„The authors' intention [...] was to illustrate [...] that the aim of developing driver models is not to create 'the' driver model [...]. Actually, the level of detail to integrate in a driver model really depends on the goal of the developer. For instance, in order to predict driver behaviour, how necessary is it to describe the psycho-motor processes underlying the driving activity or is a data analysis of data patterns and trends sufficient?"* [43], azaz „A szerzők szándéka az volt, hogy megmutassák: a vezetői modellek fejlesztésének célja nem „a" vezetői modell kifejlesztése. Egy vezetői modell részletessége nagyban függ annak alkalmazásától. Például, hogy megjósoljuk egy vezető viselkedését, mennyire szükséges leírni a vezetés mögött rejlő pszicho-motoros folyamatokat, vagy elég egy adathalmazban jelenlévő trendeket felfedni és modellezni?"[1] Azaz ismerni kell a teljes vezető, szélesebb spektrumon pedig a teljes közlekedési rendszer elemeit és azok egymásra gyakorolt hatását, majd ennek tudatában kiválasztani az alkalmazásnak megfelelő modellt. Az 1. táblázatban látható két dimenziót egymással keresztezve összesen 4 jól elkülöníthető modell osztályt kapunk, amelyek összessége adja a teljes vezetői modellt. Itt két dologra hívjuk fel a figyelmet:
- egy értelmezés szerint minden vezetési szituációban mindegyik modell osztály jelen van, és így ezek szuperponált hatása eredményezi a végső történéseket, másrészről viszont
- a fentebb említett modellezési cél szerint pontosítani kell a modell alkalmazását, és ennek megfelelően akár egyszerűsíteni annak struktúráján, így elméletileg előfordulhat az, hogy tisztán csak az egyik modell osztály összefüggéseit használjuk fel.

Ez a két szempont egymással szemben áll. A modellezési probléma megfogalmazása során mi magunk is megmutatjuk, hogy mindkét utat választhatjuk, ugyanakkor az egyszerűsítés esetén fontos olyan peremfeltételeket szabni, amely ezt az egyszerűsítést lehetővé teszik (pl. kifejezetten a kormányzással kapcsolatos mechanikai modell egy funkcionális ki- és bemeneti modell, a modellre hatással lehetnek a vezetői saját tulajdonságai, és ez így már egy tulajdonság modell is; ahhoz, hogy ezt feloldjuk, ki lehet jelölni egy választott névleges viselkedést, pl. tapasztalt sofőr).

A fenti gondolatokat figyelembe véve az alábbi követelményeket támasztjuk a vezetői modellünk felhasználási területével kapcsolatban:
- a vezetői modellt kizárólag normál aszfaltos úton, jó minőségű felületen, egy kijelölt sávon belüli haladásra használjuk, amely esetben a jármű a 60 – 110 *kph* sebeségtartományban közlekedik,
- a modell célja, hogy megadja a sávon belüli kívánt jármű pozíciót, orientációt és görbületet, nem célja a jármű sebességének tervezése,
- továbbá nem vesszük figyelembe a dinamikusan változó körülményeket, azaz más járművek, illetve közlekedési résztvevők mozgását,

---
[1] A szerző szabadszavas fordítása angol nyelvről.

- a modell lokálisan érzékelt valós idejű információk alapján képezi a lokális útvonalat.

Ezen szempontok alapján mondhatjuk, hogy a cél egy olyan vezetői modell megalkotása, amely képes emberszerű ívek tervezésére az út geometriai információi alapján. A modellt ezek után a következő alkalmazásban használjuk:
- a modell segítségével mérhetők, illetve reprodukálhatók az egyéni ívválasztási preferenciák,
- a modell segítségével osztályozható az adott vezető vezetési típusokba, és ez az információ tetszőlegesen felhasználható.

### b. A vezetői modell azonosítása

A 2/c fejezetben megállapítottuk, hogy a vezetői modellrendszerek 4 dimenzióval rendelkeznek, továbbá bevezettünk alapfogalmakat, melyek segítségével behatárolható a 3/a pontban megfogalmazott modell. A következőkben elhelyezzük az általunk modellezni kívánt szakaszt a modellrendszeren belül. A következő szcenáriót illusztráljuk először: a vezető számára ismert, jól belátható útszakasz, eldöntött útvonal (azaz nem szükséges sávváltás, kanyarodás…stb.), nincs más jármű a sávban, nincs szembejövő forgalom, a vezető ismeri az autót, kipihent és nagy vezetési tapasztalattal rendelkezik, továbbá nyugodt, kiegyensúlyozott személyiség. Ennek a helyzetnek az illusztrált modell-alrendszerét az 5. ábra (a) részábrája mutatja. A sávon belüli haladás járműszintű irányítása a manőver szinthez tartozik, így a feladatokat feketével kereteztük be ezen a szinten. Feltételezhetjük, hogy ebben a szcenárióban az észlelés elsősorban készség alapú, azaz nem tudatos a vezető részéről. Mivel nincsen más mozgó objektum a közelben, így elsősorban a vezethető útfelület észlelése a fő cél. A döntéshozás, azaz a befutni kívánt útvonal megtervezése részben készség, részben szabály alapú, hiszen ösztönösen választunk egy útvonalat, de figyelembe vesszük pl. az út geometriáját, a jármű aktuális sebességét…stb., így valamilyen belső motiváció modell szabályai szerint is döntünk. A végrehajtás feltételezhetően készségalapú, ösztönösen választunk pl. célgyorsulást vagy kormányszöget, azonban itt is megjelenhetnek a szabályszerűségek, hiszen nem mindig ragaszkodunk egy bizonyos útvonalhoz (amit korábban megterveztünk), hanem ettől eltérünk egy megadott tűréshatáron belül, belső szabályaink szerint („*risk threshold theory*"). A teljesség kedvéért felrajzoltuk az operatív szinten zajló feltett folyamatokat, amelyet mind a készségalapú megvalósításba soroltunk, mind az észlelés, mind a döntéshozás, mind a végrehajtás terén (feltettük, hogy a vezető rendelkezik annyi vezetői tapasztalattal, hogy pl. a kormányszög megvalósítás, vagy a sebesség tartása ne okozzon neki különösebb tudatos megterhelést).

|  |  | Észlelés | Döntéshozás | Végrehajtás |
|---|---|---|---|---|
| Stratégiai | Tudásalapú |  |  |  |
|  | Szabályalapú |  |  |  |
|  | Készségalapú |  |  |  |
| Manőver (taktikai) | Tudásalapú |  |  |  |
|  | Szabályalapú |  | 🟩 | 🟩 |
|  | Készségalapú | 🟩 | 🟥 | 🟩 |
| Szabályzó (operatív) | Tudásalapú |  |  |  |
|  | Szabályalapú |  |  |  |
|  | Készségalapú | 🟢 | 🟢 | 🟢 |

|  |  | Észlelés | Döntéshozás | Végrehajtás |
|---|---|---|---|---|
| Stratégiai | Tudásalapú |  |  |  |
|  | Szabályalapú |  |  |  |
|  | Készségalapú |  |  |  |
| Manőver (taktikai) | Tudásalapú |  |  |  |
|  | Szabályalapú |  | 🟥 | 🟦 |
|  | Készségalapú | 🟦 | 🟥 | 🟦 |
| Szabályzó (operatív) | Tudásalapú |  |  |  |
|  | Szabályalapú |  |  |  |
|  | Készségalapú | 🟠 | 🟠 | 🟠 |

**(a)** **(b)**

**5. ábra. Az 1. szcenárió (a) és a 2. szcenárió (b) illusztrált modellje a 4-dimenziós modellrendszerben elhelyezve**

A magyarázat kedvéért illusztrálunk egy másik szcenáriót is. Ez így hangzik: reggeli csúcsforgalom, sávban haladás, kipihent, ám sietős vezető, napos idő, ismert útszakasz, ismert jármű, a sávunkban jelenlévő más járművek, országút, szembejövő járművek. Ennek a feltételezett modell alrendszerét az 5. ábra (b) alábrája jelöli.

Ebben az esetben is a sávon belüli ívválasztás a manőverezés szintjére esik. Az észlelés feltételezhetően továbbra is készségalapú, a döntéshozás (azaz az ív megtervezése) inkább szabály alapú, hiszen sok a külső változó, így a belső szabályrendszer szempontjai felerősödnek. A végrehajtás inkább készség alapú, hiszen a változók miatt feltételezhetően nagyobb a koncentráció, így az eltervezett ívet pontosabban hajtja végre a sofőr.

Az eddigiekben nem esett szó az egyes cellákban szereplő színekről. Az 5. ábra táblázata tartalmazza a Theeuwes féle 3 dimenziót, azonban nem tér ki a 4-re, a viselkedési modellre. Ahogyan azt a 3/a pontban is megadtuk, az útvonal tervező vezetői modell segítségével szeretnénk megfigyelni és reprodukálni egyéni ívválasztásokat. Így az alkalmazásunkra nézve is fontos, hogy a viselkedési modell szerepeljen a modell rendszerben. Ezt a 4. dimenziót jelképezik a színek. Megjegyezzük, hogy a választott szín csupán illusztráció. A szín jól megtestesíti a viselkedés azon tulajdonságát, hogy a viselkedés nem diszkretizálható mennyiség, továbbá önmagában is komplex modellt alkot. Ahogy egy szín lehet különböző árnyalatú, fényerősségű, tágabb értelemben akár különböző textúrájú, úgy a viselkedési minták is szerte ágazók. Az 1. szcenárió esetén zöld szint használunk mindenhol, ezzel reprezentálva a nyugodt, kiegyensúlyozott viselkedést. Ahol a szín erősebb, ott a modell tulajdonságok intenzíven játszanak szerepet a modellrendszerben (pl. erős szín a készség és szabályalapú megközelítések határon, halványabb ettől messzebb), ahol halvány, ott pedig kevésbé játszik szerepet. Ahol nem szerepel szín, ott úgy véljük, az adott szcenárióban nem releváns az az modellrész, így nem befolyásolja a végső kimenetet. A 2. szcenárió esetén a kék szín az odafigyelést, a sietésből adódó intenzív észlelést jelöli, míg a piros a dinamikus, agresszív tulajdonságokat testesíti meg. Valamivel halványabb piros jelöli az operatív szint feladatait, jelezve, hogy éppen a sietés, kapkodás miatt kevesebb figyelem hárul a végrehajtásra.
Véleményünk szerint a fenti modell segítségével a legtöbb szcenárió esetén elhelyezhető a modellezni kívánt szakasz.

A 3/a pontban megfogalmazott alkalmazásnak megfelel az 1. szcenárió. A modellt kizárólag a manőverezési feladat döntéshozási szintjére készítettük el. Ez a szint a kiválasztott ív megtervezését jelenti lokálisan, kizárólag statikus információk alapján (azaz más járművek, illetve közlekedési szereplők viselkedését nem vesszük figyelembe). Továbbá ennek a modellnek a célja egy ösztönös viselkedés leírása, azaz kizárólag a készségalapú megközelítést használjuk. A szabályalapú megközelítés esetén figyelembe vehetnénk több tényezőt, ezáltal egy optimalizációs problémát kapunk. A modell elkészítésénél azonban ezt nem vettük figyelembe, ehelyett feltártunk egy fizikai modellt, mely az alapját adja a szabályalapú megközelítésnek. A modell helyét a modellrendszerben az 5. ábra (a) alábráján piros téglalap jelöli.
A modell azonosításához valódi sofőrök mért adatait használtuk. A modell bemenetei az út belátott szakaszának geometriai viszonyai, a kimenete pedig egy lokális útvonal, amely tükrözi az adott sofőr preferenciáit. A viselkedési modellt az útvonaltervező modell paraméterei reprezentálják. Ezeket a paramétereket külön-külön megfigyeltük az egyes vezetőkre.

Biztosítottuk, hogy mindegyik vezető ugyanolyan körülmények között, nyugodt lelki állapotban teljesítse a vezetést. Az időbeli paraméterváltozással, azaz a viselkedési minták változásával nem foglalkoztunk. A viselkedési modell így tekinthető egy egyszerűsített modellnek.

Összefoglalva, a modellezés során a következő szempontokat vettük figyelembe:
- a vezető ösztönös, készség alapú útvonalválasztása,
- nyugodt, ideális körülmények közötti vezetés,
- országúti útszakasz, más járművek mozgásának figyelmen kívül hagyása,
- a manőverezési szint észlelésének és végrehajtásának figyelmen kívül hagyása.

Azon listázott szempontok, melyeket nem vettünk figyelembe, zavarként jelennek meg a rendszerben. Ugyanakkor a cél nem a teljes viselkedés reprodukálása, sokkal inkább az ösztönös és szándékos útvonal választási preferenciák feltárása. Az így megszületett modellt korábbi cikkeinkben mutattuk be [25], [44], [45]. A modell neve *Extended Linear Driver Model* (ELDM). A modell egy 150 méteres előre tekintési távon látott útgörbület alapján 3 csomópontban meghatározza a sávközéphez viszonyított célofszetet, majd ezen csomópontokra illeszt 3 Euler-görbét, így kiadva a teljes lokális útvonalat. A tervezés ciklikusan ismétlődik, ahogy a jármű halad. Az ofszet értékek a csomópontok közötti átlagos útgörbület lineáris kombinációjaként áll elő. A 3 görbületérték leképezése 3 ofszet értékké így összesen 9 paramétert jelent. Mivel más a vezetők jellemző viselkedése bal- illetve jobbkanyarban, így a görbület előjelének a függvényében a modell más paraméterek alapján számítja a célofszet értékeket. Ezzel nem 9, hanem 18 paramétert tartalmaz a modell. Továbbá az egyenesekben vett ofszet preferenciát egy paraméterrel adjuk meg (az átlagos ofszet értéke zérus görbület mellett), ezzel 19 paramétert kapunk. A 19 paramétert minden egyes vezető nagyjából 40 km-es vezetése után határoztuk meg, külön-külön.

Ahhoz, hogy ne csak megfigyelni, de reprodukálni is tudjuk az egyéni preferenciákat, több megoldás is rendelkezésre áll. Egyrészt kiszámíthatjuk az egyéni paramétereket, majd pl. egy sávtartó aktiválásánál az adott sofőrre jellemző paraméterhalmazt használjuk. Ezt az esetet a 6. ábra (a) alábrája mutatja. Azonban ebben az esetben a végleges kódban szabad paramétereket hagyunk, ami így megnehezíti a verifikációt és validációt. Sorozatgyártásba szánt szoftver esetén jobb megoldás, ha a vezetőket a saját viselkedésünk alapján típusokba soroljuk, majd az adott típusú sofőrökre jellemző viselkedést reprodukáljuk az egyéni viselkedések helyett. Ezt a verziót a 6. ábra (b) alábrája mutatja. A továbbiakban ezt a megvalósítást fogjuk használni.

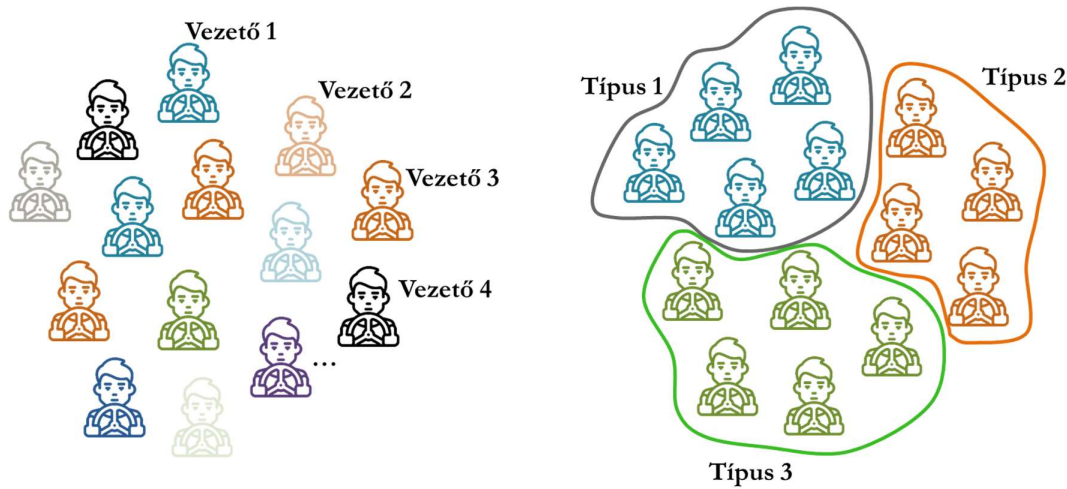

**6. ábra.** A vezetők osztályozása az ívválasztásuk alapján, (a) egyéni viselkedések reprodukálása, (b) vezetői típusok használata

A tanítási adathalmazban összesen 19 sofőr szerepelt. Minden egyes sofőrt 19 paraméterrel jellemeztünk, majd ezt a 19 paramétervektort használtuk fel a vezetők osztályozásához. A következőkben ezen klaszterezési eljárást mutatjuk be.

## 4. Vezetők klaszterezése vezetői típusokba

Az ELDM modell alapján elkülöníthető vezetési típusok meghatározásához k-means [46], illetve hierarchikus [47] klaszterezési módszereket használtunk fel. Sűrűség alapú klaszterezés használatát az adatkészletünk dimenziója, illetve a minták darabszáma miatt vetettük el. A klaszterezéshez használt adatkészletünk 19 mintát tartalmazott, amely 15 vezető által megtett 40 km-es országúti útszakaszból kinyert modellek $P_{left}, P_{right}, \Delta y_0$ paramétereit tartalmazta egy $x \in \mathbb{R}^{19 \times 1}$ vektor formájában. A modell $\Delta y_0$ statikus oldalirányú eltolását meghatározó paraméterét egyszer tartalmazta egy minta a jobb eredmények elérésének érdekében. A k-means klaszterezés esetén a K paraméter meghatározására kétféle módszert alkalmaztunk. Először az ún. könyök módszer („elbow method") [48] alapján szerettük volna meghatározni K értéket, azonban az inercia-K függvényen nem volt egyik K érték esetében se kiemelkedő töréspont, ahogy azt a 7. ábra is mutatja, ezért más módszert alkalmaztunk.

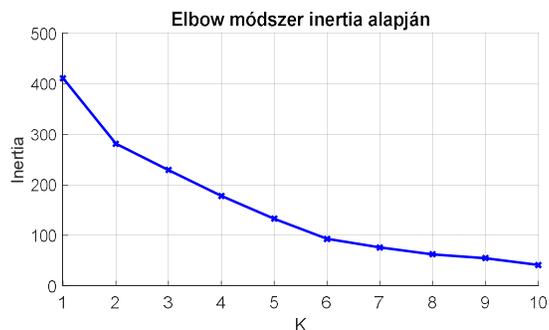

**7. ábra.** Inertia értékek a K paraméter függvényében

Az ún. *Silhouette* [49] értékeket kiszámolva az adatkészletünk egyes mintáira és ezek átlagát véve meg tudtuk határozni az optimális *K* értéket *Euklédeszi*, illetve *Manhattan* távolság becslés esetében. A korábban említett távolság becslő módszer esetében *K=3* bizonyult a megfelelő értéknek, mert a keletkezett klasztereknek legalább 1 mintája egyértelműen meghaladta az kimenet átlagos *Silhouette* értékét. Az eredményeket a 8. ábra mutatja.

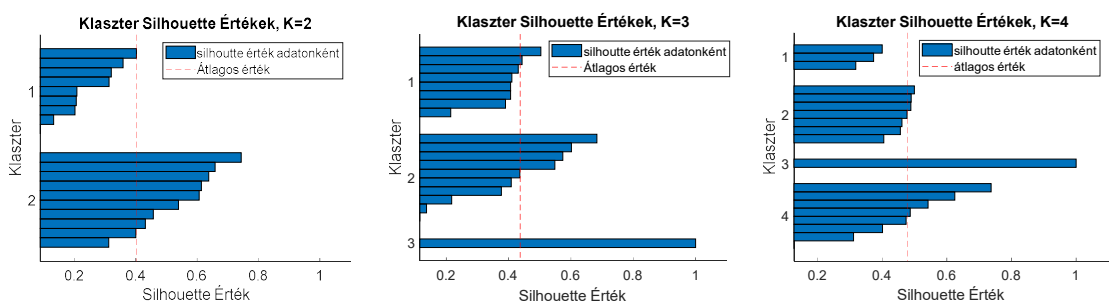

**8. ábra. Silhouette értékek K=2/3/4 paraméter esetén, Euklédeszi távolságbecslést használva**

A Manhattan távolság becslő használata esetében *K=2* bizonyult az optimális értéknek, mivel a klaszterezett mintákra számolt *Silhouette* értékek mind pozitívak, illetve a korábbi kritériumnak is megfelelt a klaszterezés eredménye. Ezeket az eredményeket a 9. ábra mutatja.

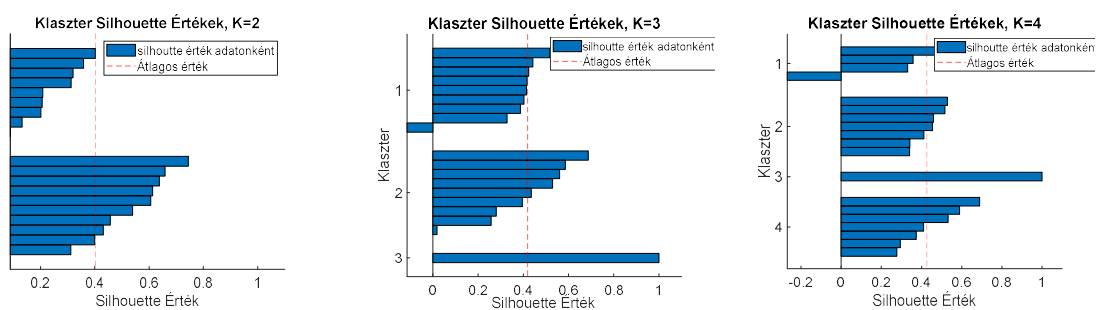

**9. ábra. Silhouette értékek K=2,3,4 paraméter esetén, Manhattan távolság becslés**

Az agglomeratív hierarchikus klaszterezés esetében a *Cophenet* korreláció [50] értéket alkalmaztuk az adatunkhoz legjobban illeszkedő távolság becslő kiválasztására, ami alapján a Manhattan távolság becslő bizonyult optimálisnak. A távolságok számítása a klaszterek között az (1) alapján történik.

$$d(r,s) = \frac{1}{n_r n_s} \sum_{i=1}^{n_r} \sum_{j=1}^{n_s} dist(x_{ri}, x_{si}) \tag{1}$$

A klaszterezés eredményét a 10. ábra mutatja. A klaszterek határértékeit 15 és 20 közé becsültük, mivel ez még nem haladja meg a maximális klaszter távolság felét. A határértéken felül eső klaszterek átlagos távolsága nagy, ezért ezek tekinthetőek azon vezetők csoportjának, amelyek nem sorolhatóak be egyértelműen egyik típushoz se.

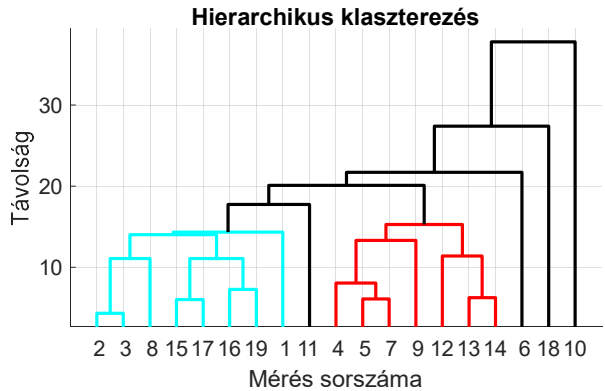

**10. ábra. A hierarchikus klaszterezés dendrogramja**

A korábban említett klaszteretési eljárások alapján kapott eredményeket a 2. táblázat mutatja. Ez adja meg, mely klaszterekbe kerültek az egyes vezetők. A *K-means* esetében az *Eukledeszi* távolság becsléssel kapott eredményeket ábrázoltuk, mivel ezek a klaszterek nagyobb átfedésben voltak a hierarchikus klaszterezés eredményével.

| Vezetők | 1/1 | 1/2 | 2/1 | 2/2 | 2/3 | 3 | 4/1 | 4/2 | 5 | 6 | 7 | 8 | 9 | 10 | 11 | 12 | 13 | 14 | 15 |
|---|---|---|---|---|---|---|---|---|---|---|---|---|---|---|---|---|---|---|---|
| K-means | 1 | 1 | 1 | 3 | 1 | 1 | 3 | 3 | 3 | 1 | 3 | 3 | 1 | 1 | 3 | 2 | 1 | 3 | 1 |
| Hierarchikus | 1 | 1 | 1 | 3 | 2 | 1 | 3 | 3 | 3 | 1 | 2 | 3 | 1 | 1 | 3 | 2 | 2 | 3 | 1 |

**2. táblázat. Az adatkészleten belül kialakult klaszterek minták alapján**

Az egyes klaszterekre jellemző modell paraméter értékeket a klaszterbe tartozó minták átlagaként határoztuk meg, ami egyben a klaszterek középpontjait is jelenti. Ezeket az átlagos paraméter értékeket (2) és (3) adja meg. Egy újabb vezető megfigyelése után eldönthető, melyik vezetési klaszterbe sorolható ez a vezető.

$$P_1 = \begin{bmatrix} 1.26 & -1.32 & 0.14 & 0.16 & 0.03 & -0.12 & -0.09 \\ 0.37 & -0.24 & -0.05 & -0.75 & 1.07 & -0.25 & -0.09 \\ -0.60 & 0.69 & 0.00 & -0.35 & 0.21 & 0.22 & -0.09 \end{bmatrix} \quad (2)$$

$$P_3 = \begin{bmatrix} -0.45 & 0.64 & -0.21 & -0.38 & 0.71 & -0.26 & 0.00 \\ -1.00 & 1.32 & -0.33 & -1.67 & 2.21 & -0.48 & 0.00 \\ -0.51 & 0.66 & -0.17 & 0.58 & -0.78 & 0.28 & 0.00 \end{bmatrix} \quad (3)$$

A 2-es klaszterhez nem határoztunk meg jellemző paramétert, mert a *K-means* módszer esetén a klaszter minták eloszlása alapján, a hierarchikus klaszterezés esetén pedig a dendrogram alapján látszik, hogy azok a minták egy kívülálló csoportot, mint sem egy típust határoznak meg. Az 1-es és 3-as számú klaszterek a klaszterezés típusától függetlenül átfedésben vannak egymással, ezzel is bizonyítva az összetartozásukat.

## 5. Eredmények

A 4. fejezetben bemutatott klaszterezési eljárás alapján elkülönítettük az egyes vezetőket, illetve meghatároztuk a klaszterek jellemző viselkedését, mint a klaszterek átlagos paramétereit. Ezekkel szimulációt végeztünk egy rövidebb kanyar kombinációban. Ezt az útszakaszt a 11. ábra mutatja. Ez a 31-es út Mende és Maglód közötti szakasza. A járművet a szimulációban egy kinematikai biciklimodell helyettesíti, a szabályzó egy pure-pursuit szabályzó. Mivel a járműmodell egyszerűsített, így a szabályzó közel zérus pozíció hibával tudja lekövetni a kívánt útvonalat. A vizsgált szakasz egy enyhe jobb kanyarral kezdődik, majd egyenesen halad. Ezek után következik egy éles S-kanyar, először bal majd jobb kanyarokkal. Egy nagyon rövid szakasz után újabb S-kanyart látunk, amely először jobbra majd balra ível, azonban kisebb görbülettel, mint az első S-kanyar. Végezettül újra egyenes szakasz következik. A 11. ábra felső diagramja mutatja az útvonalat a globális UTF koordinátarendszerben, fekete körrel jelölve a jármű kiinduló pontját. Az alsó diagram az út görbületét mutatja: pozitív görbület felel meg a bal kanyarnak.

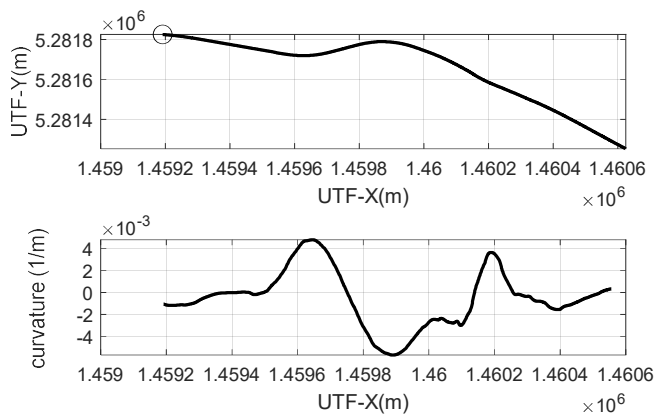

**11. ábra. A validációs útszakasz, amit a szimulációhoz használtunk**

Először a 2. táblázat első sorában látható K-means klaszterezési eredményeket felhasználva szimuláljuk az 1-es és 3-as sz. klaszterhez tartozó viselkedéseket. A teljesség kedvéért az átlagos viselkedés mellett az egyéni viselkedéseket is megjelenítjük. Az eredményeket a 12. ábra mutatja. A következő megállapításokat tehetjük:

- 1. sz. marker: az 1. klaszter egyenesbeli ofszete negatív.
- 2. sz. marker: közel szimmetrikus viselkedés, maximális ofszet: ~ +/- 1 $m$.
- 3. sz. marker: első ofszet csúcspont később található a megtett táv szerint, mint a 3. klaszter esetén, azaz az 1. klaszterben szereplő vezetők általánosságban a kanyarban később húzódnak át a belső ívre.
- 4. sz. marker: első bal kanyar előtt negatív ofszet, amely egyúttal ív külsőre húzódást is jelent. Ez megfigyelhető a 3. klaszter esetében is, azonban a csúcspont előrébb van (hasonlóan a 3. sz. markerhez).
- 5. sz. marker: hosszan tartó pozitív ofszet a balkanyarban (kanyarlevágás). Ugyanez a 3. klaszter esetén nem figyelhető meg (sávközépre húzódás). Ez egyben a jobb kanyarra való készülés az 1. klaszter esetén, a 3. klaszter esetén ez a felkészülés jól elkülöníthető.
- 6. sz. marker: jobb kanyarban a levágás mértéke és hossza közel azonos mindkét klaszter esetében. Az 1. klaszter esetén ez elnyúlik, és egyben az utolsó balkanyarra való készülést is jelenti. A 3. klaszter esetén ez jól elkülöníthető (7. sz. marker).
- 8. sz. marker: az utolsó, jobb kanyar levágása hasonló mértékű és hosszúságú mindkét klaszter esetén, azonban a 3. klaszter esetén a csúcspont hamarabb van.
- 9. sz. marker: aszimmetrikus maximális ofszet értékek bal- és jobb kanyarban: ~+0.75 $m$ / -1 $m$.
- 10. sz. marker: a 3. klaszter esetén közel nulla ofszet egyenesekben.

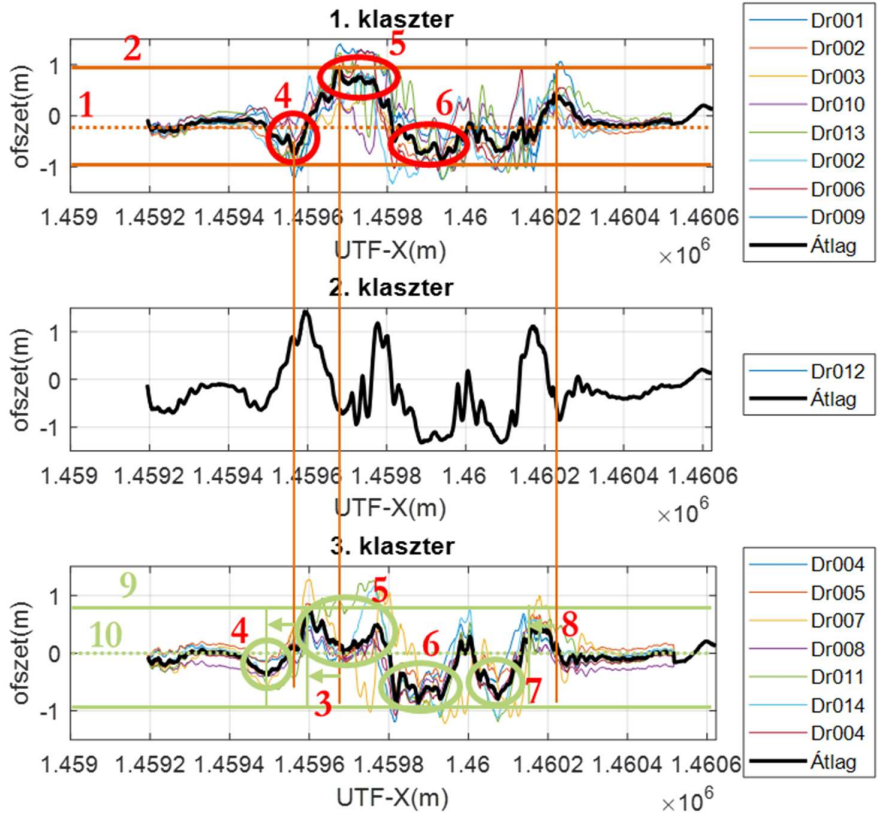

**12. ábra. A K-means klaszterező alapján meghatározott vezetési típusok ívválasztása a szimulált kanyarkombinációban.**

Ugyanígy elvégeztük a szimulációt a 2. táblázat 2. sorának eredményei alapján, ahol a hierarchikus klaszterezési módszert használtuk. Az eredményeket a 13. ábra mutatja. Itt is hasonló jelenségeket figyelhetünk meg, mint a *K-means* alapú klaszterezés esetén:

- 1. sz. marker:
    - Egyenesbeli ofszet negatív (1. klaszter)
    - Nem egyértelmű, inkább negatív (2. klaszter)
    - Közel nulla (3. klaszter).
- 2. sz. marker:
    - Közel szimmetrikus: +/- 0.9 *m* (1. klaszter)
    - Közel szimmetrikus: +/- 1.2 m (2. klaszter)
    - Aszimmetrikus: + 0.65 *m* / - 1 *m* (3. klaszter)
- 3. sz. marker:
    - Első csúcspont kanyarban későn (1. klaszter)
    - Első csúcspont a kanyar előtt (2. és 3. klaszter)
- 4. sz. marker:
    - Kanyar előtti külső ívre húzódás nagy mértékű (~0.8 *m*), kanyarbejáraton (1. klaszter)
    - Kanyar előtti külső ívre húzódás kisebb mértékű (~0.3 *m*) és a kanyar előtt található (2. és 3. klaszter)

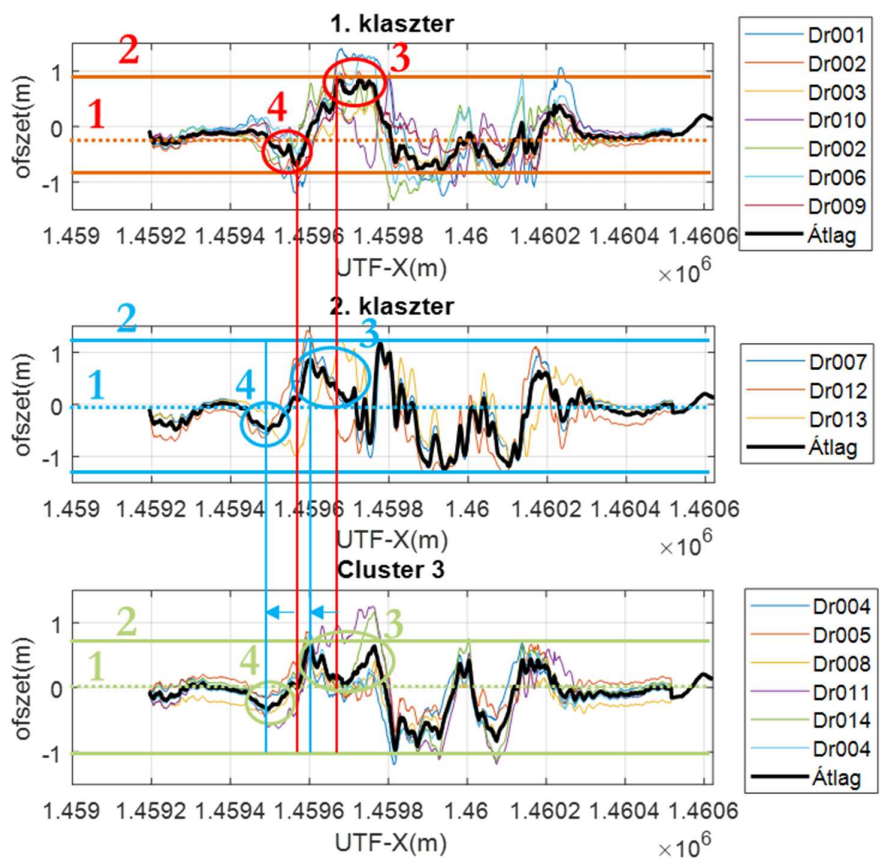

**13. ábra. A hierarchikus klaszterező alapján meghatározott vezetési típusok ívválasztása a szimulált kanyarkombinációban.**

Megjegyezzük, hogy míg a *K-means* esetén a 2. klaszter csupán egy kívülálló elemet tartalmaz, addig a hierarchikus klaszterezés esetén ez már 3 vezetőt jelent, így itt külön értékeltük ezt a klasztert is. Azonban a megfigyelések alapján nem eldönthető, hogy a 2. klaszter rendelkezik e saját tulajdonságokkal, sokkal inkább egyfajta keveréke az 1. és a 3. klaszter viselkedési mintáinak. Ebből azt a következtetést vonjuk le, hogy a 2. sz. klaszter valóban kívülálló elemek csoportja, és ezen az adathalmazon nem eldönthetők egyértelműen a jellemzőik.

|  | Vezetői típus 1. | Vezetői típus 2. |
|---|---|---|
| **Az abszolút jobb- és balkanyarbeli maximális ofszet helye és értéke** | helye: a kanyarban később értéke: szimmetrikus, +/- 1 $m$ | helye: a kanyar bejáratához közel értéke: aszimmetrikus, +0.75 $m$ / -1 $m$ |
| **Kanyar előtti ívkülsőre húzódás helye és mértéke** | helye: közel a kanyarhoz mértéke: erőteljes | helye: kanyar előtt hamarabb mértéke: nem jelentős |
| **Kanyarlevágás helye és hossza** | helye: a kanyarban később hossza: hosszan elnyúlóan | helye: kanyar bejáratán hossza: rövid ideig |
| **Egyenesbeli ofszet mértéke** | negatív | közel nulla |

**3. táblázat. A vezetői típusokra jellemző viselkedések a feltárt 4 db kanyarbeli jellemző szerint**

Az eredményeket a 3. táblázat mutatja. Sikerült bebizonyítani, hogy az egyes klaszterekbe sorolt vezetők hasonló tulajdonságokkal rendelkeznek, ezek a tulajdonságok 4 jellemző alapján megfogalmazhatók: abszolút jobb- és balkanyarbeli maximális ofszet helye és értéke, kanyar előtti ívkülsőre húzódás helye és mértéke, kanyar előtti ívkülsőre húzódás helye és mértéke és egyenesbeli ofszet mértéke. Ezek alapján az elkülönített 2 vezetői klasztert nevezhetjük vezetői típusnak. Ahogyan azt a 4. fejezetben is említettük, bármely új vezető megfigyelése után eldönthető, hogy az adott vezető átlagos ívválasztási viselkedése melyik típusba sorolható. Ezek után az adott típushoz tartozó jellemző viselkedést reprodukálva emberszerűvé tehetjük a sávtartás funkciót.

## 6. Konklúzió

Ebben a cikkben bemutattuk az általános vezetői modellezési problémát, részletes áttekintést adtunk a történelméről és felhasználva a legújabb modellrendszereket egy megfogalmaztuk a sávtartás problémájához tartozó modellezési feladatot. Ezen kívül felhasználva egy korábban alkotott ívválasztási modellt sikerült valódi vezetők preferenciáit felismerni, illetve ezek alapján a vezetőket klaszterekbe sorolni. A klaszterekhez megadtunk egy átlagos viselkedést, amelyhez összesen 4 jellemző alapján társítottunk tulajdonságokat. Ezáltal sikerült 2 különálló, saját viselkedéssel bíró vezetői típust elkülöníteni. A meglévő adathalmaz egyértelműen nem alkalmas arra, hogy eldöntsük, elegendő a 2 vezetői típus az ívválasztás alapján, ugyanakkor állíthatjuk, hogy legalább 2 típus létezik. Ezzel alkalmazást adtunk az *Extended Linear Driver Modelnek*, és ennek segítségével egy sávtartó rendszer személyre szabható, emberszerűvé tehető.

A jövőben szeretnénk az 5. ábra szerinti 4 dimenziós vezetői modellt tovább vizsgálni kifejezetten a sávtartás problémájánál maradva. Úgy gondoljuk, hogy az ívtervezésben nagy szerepet játszanak a külső zavarok, amelyekre adott szabályszerű válaszreakciót szeretnénk modellezni. ezáltal szélesítve a modell felhasználási körét.

## Irodalomjegyzék